\documentclass[conference]{IEEEtran}
\IEEEoverridecommandlockouts
\usepackage{cite}
\usepackage{amsmath,amssymb,amsfonts}
\usepackage{algorithm}
\usepackage{algorithmic}
\usepackage{graphicx}
\usepackage{textcomp}
\newcommand{\RomanNumeralCaps}[1]{\MakeUppercase{\romannumeral #1}}
\newcommand{\quotes}[1]{``#1''}

\usepackage{xcolor}

\DeclareMathOperator*{\argmax}{arg\,max}
\def\BibTeX{{\rm B\kern-.05em{\sc i\kern-.025em b}\kern-.08em
    T\kern-.1667em\lower.7ex\hbox{E}\kern-.125emX}}
    
\graphicspath{{./figs/}}

\begin{document}

\title{\textsc{GraMeR}:Graph Meta Reinforcement Learning for Multi-Objective Influence Maximization}

\author{Sai Munikoti$^{1}$, Balasubramaniam Natarajan$^{1}$ and Mahantesh Halappanavar$^{2}$
\thanks{$^{1}$S. Munikoti and B. Natarajan are with Electrical and Computer Engineering, Kansas State University, Manhattan, KS-66506, USA.
        {\tt\small saimunikoti@ksu.edu, bala@ksu.edu}}%
\thanks{$^{2}$M. Halappanavar is with Data Science and Machine Intelligence group, PNNL, Richland, USA. 
        {\tt\small mahantesh.halappanavar@pnnl.gov}}
	\thanks{This work has been submitted to the IEEE for possible publication. Copyright may be transferred without notice, after which this version may no longer be accessible.}%
}

\maketitle

\begin{abstract}
Influence maximization (IM) is a combinatorial problem of identifying a subset of nodes called the seed nodes in a network (graph), which when activated, provide a maximal spread of influence in the network for a given diffusion model and a budget for seed set size. IM has numerous applications such as viral marketing, epidemic control, sensor placement and other network-related tasks. However, the uses are limited due to the computational complexity of current algorithms. Recently, learning heuristics for IM have been explored to ease the computational burden. However, there are serious limitations in current approaches such as: (1) IM formulations only consider influence via spread and ignore self activation; (2) scalability to large graphs; (3) generalizability across graph families; (4) low computational efficiency with a large running time to identify seed sets for every test network. In this work, we address each of these limitations through a unique approach that involves (1) formulating a generic IM problem as a Markov decision process that handles both intrinsic and influence activations; (2) employing double Q learning to estimate seed nodes; (3) ensuring scalability via sub-graph based representations; and (4) incorporating generalizability via meta-learning across graph families. Extensive experiments are carried out in various standard networks to validate performance of the proposed Graph Meta Reinforcement learning (GraMeR) framework. The results indicate that GraMeR is multiple orders faster and generic than conventional approaches.
\end{abstract}

\begin{IEEEkeywords}
graph neural networks, Q learning, influence maximization
\end{IEEEkeywords}

\section{Introduction}
Consider a directed graph $G=(V, E, \omega)$, where $V$ is a set of vertices, $E$ is a set of edges (pairwise relationships on vertices), and $\omega$ is a set of edge weights; a diffusion model, and a budget $k$.
Influence maximization (IM) is the problem of identifying a set of $k$ seed nodes, which when activated, will result in the activation of a maximal number of nodes in $G$, for the given diffusion model of influence. IM has applications in various domains ranging from viral marketing in social networks to influential proteins in biological networks. For instance, one of the reasons behind the tremendous success of social media platforms is the quality of content the users create or generate via sharing. These actions can be attributed to influence dynamics in the social network. IM was first introduced in 2001 and was formulated as a combinatorial optimization problem
by \cite{kempe2003maximizing}. Majority of the IM algorithms
focus on settings where seed nodes are activated deterministically and then neighborhood nodes are activated via influence. However, there exists two types of activation in real-life scenarios, namely intrinsic and influenced \cite{sathanur2018exploring}. In case of social media, intrinsic refers to users who post content whereas sharing, retweeting, commenting constitute influenced activation. To this end,
the authors in \cite{farajtabar2014shaping} recognize that the events on social media can be categorized as exogenous and endogenous and model the overall diffusion through a multivariate Hawke’s process to address activity shaping in social networks. In another recent work \cite{quach2016diffusion}, the authors propose a novel diffusion model based on factor graphs and graphical models where the node potentials can correspond to the notion of intrinsic activation. However, the focus of their work is on the diffusion model itself, not on the aspects of intrinsic activation.

Furthermore, despite the huge potential of IM, its field uses are limited due to the computational inefficiency of conventional greedy algorithms. Therefore, recently, there have been studies that use Reinforcement learning (RL) to learn generalized policy for combinatorial optimization
problems on graphs \cite{dai2017learning,li2018combinatorial,hu2020reinforcement} including IM. The key idea is
to decompose the node selection process into a sequence, and
learn a heuristic policy that selects nodes sequentially. The
RL policy is usually trained on a set of seen training graphs,
in the hope that it generalizes to unseen test graphs of similar characteristics. To better generalize the trained policy
across different graphs, graph embedding techniques, such
as Structure to Vector (S2V)\cite{dai2017learning} and Graph
Convolutional Networks (GCNs) \cite{kipf2016semi}
are integrated as part of the RL value functions to extract
the graph structure information. Primarily proposed to solve
relatively simple tasks such as the traveling salesman
problem (TSP) and the maximum vertex cover (MVC), some of the recent works \cite{manchanda2020gcomb, chen2021contingency} extend it to the IM problem. However, they possess following shortcomings: 
\textbf{ (1) Formulation}: Existing works such as GCOMB, S2V-DQN and GCN-TREESEARCH addresses a trivial IM problem without explicitly accounting the intrinsic and influence activations. Thus, these methods could not work effectively in real-world scenarios.
\textbf{ (2) Computational complexity}: Most of the conventional IM works such as CHANGE and \cite{sathanur2018exploring} are computationally complex and thus possess limitations in transitioning solution to real-life uses. These methods require re-computation of IM nodes whenever a underlying network changes and usually the stakeholders do not have the HPCs to perform repetitive computation.
\textbf{(3) Scalability}: To counter computational complexity of existing IM algorithms, RL based learning algorithms have been recently proposed in the literature. The primary focus of these methods (S2V-DQN and RL4IM) are on obtaining high quality results. Howevever, the efficiency studies are limited to graphs containing only hundreds of thousands of nodes. Real-life graphs may contain millions of nodes/edges.  
 \textbf{ (4) Generalizability}: Existing methods such as S2V-DQN, GCOMB and RL4IM does not systematically account for graph variation, i.e., trained RL model would work only on graphs of similar characteristics. 
\subsection{Contributions}
To address some of the above-mentioned shortcomings, this article proposes a novel Graph Meta Reinforcement learning (GraMeR) framework for influence maximization problem that affords both computational advantage and scalability to graphs of different sizes and families. Particularly, the article makes the contributions to the literature in the following aspects:

\begin{enumerate}
    \item The underlying IM \textbf{formulation} in this work is a more generic and realistic  in terms of incorporating intrinsic and influence activation via probabilistic paradigm.
    \item A novel GNN based approach to prune the search space while inspecting for the most influential nodes. This improves the \textbf{computational efficiency} of the proposed framework. 
    \item The RL algorithm of our framework, i.e., double deep Q learning learns the topological patterns which are recurring at every step of the Graph optimization problem. Therefore, once the model is trained, it can predict the appropriate sequence for a new graph in no time. 
    \item Meta learning is achieved by feeding the input graph information along with the state vector while estimating the Q value (long term benefit) of feasible actions. This induces \textbf{generalization} in our framework
    by allowing predictions across graphs of different families.
    \item Integration of intrinsic and influence activation requires \textbf{multi-objective RL formulation}. A single policy multi Q learning approach is implemented to learn optimal policy for a given preferences of objectives. 
\end{enumerate}
The rest of the paper is organized as follows. Section \RomanNumeralCaps{2} offers the background of IM. Section \RomanNumeralCaps{3} describes the generic IM formulation followed by proposed candidate node predictor module in Section \RomanNumeralCaps{4}. Section\RomanNumeralCaps {5} presents the main DRL framework with experiments and results in Section \RomanNumeralCaps{6}. Final conclusions are provided in Section \RomanNumeralCaps{7}.

\section{Background and Related work}
\cite{Domingos2001} and \cite{kempe2003maximizing} constitute the seminal work that introduced Influence Maximization (IM) and provided a practical approach to solve it. This work was followed by several other works that explored various diffusion models and variations to the one discussed in \cite{kempe2003maximizing} namely, the independent cascade (IC) model and the linear threshold model. 
For instance, in \cite{gionis2013opinion}, the authors consider the task of identifying the individuals whose strong positive opinion
about a product will maximize the overall positive opinion about the product. In the process, the authors leverage the social influence model proposed by Friedkin and Johnsen \cite{friendkin1999social}. 
In recent works \cite{quach2016diffusion, farajtabar2014shaping}, the authors recognize several types of activations in social network and propose methods to identify seed nodes under these activations. 

Although the IM literature has become mature enough to tackle various real-world network scenarios, there exists a serious limitation attributed to the  high computational costs in various applications \cite{srivastava2019network, rice2020using, Minutoli2019}. 
Therefore, AI/ML community focused on developing algorithms for such computationally exhaustive tasks ranging from node identification tasks to combinatorial optimization on graphs \cite{munikoti2022scalable, hussain2021influence, munikoti2021bayesian }. 
The existing literature for combinatorial and graph problems can be broadly divided into two types: First, algorithms that pose combinatorial optimization into a sequential decision-making process and learn heuristics to support the primary optimizer \cite{vinyals2015pointer, graves2016hybrid}. The application includes graph discovery in social network, restoration routes in critical infrastructure networks among others \cite{ravazzi2021learning, munikoti2021robustness}.
Major drawbacks of these approaches are generalizability and data inefficiency. Therefore, the second category of work focuses on novel graph machine learning algorithms to tackle these shortcomings. Particularly, reinforcement learning (RL) can be used for solving graph optimization problems. However, functional approximators in conventional RL algorithms are not powerful enough to learn complex tasks. 
The advent of deep learning significantly boosts the approximation capability and enables RL to solve several tasks that were not possible before. \cite{kool2018attention} is one of the pioneering works in deep reinforcement learning (DRL), which combines attention-based function approximator with policy gradient Deep RL algorithms. Further, graph convolutional networks are more appropriate for capturing topological information in graphs. In this regard, \cite{li2018combinatorial} approximate solution with graph convolutional networks (GCN) embeddings, and uses a learning framework based on guided tree search to approximate DRL solutions. Moreover, there are some works where RL is combined with GCN to solve the traveling sales man problem \cite{hu2020reinforcement}. 
Similarly, authors in \cite{wang2020grl}, solved the widely known Knowledge Graph (KG) completion problem with DRL. Particularly, they employed GCN to learn a representation of KG and then use a deep deterministic policy gradient algorithm to estimate missing relationships.

Specific to IM, Li et al. and Tian et al. are a few efforts to solve the IM problem using deep Q learning. Here, reward corresponds to marginal gain in influence spread, and agent learns to find a node set that has maximum spread. Manchanda extended the method in Khalil by solving IM problem for very large graphs \cite{manchanda2020gcomb}. Specifically, it has a separate module to predict node quality which is fed as an input to the DRL agent. This induces computational burden in the training pipeline and the method of predicting node quality is not systematic. 
Recently, \cite{chen2021contingency} developed a DRL algorithm for realistic IM by considering node willingness to be seed node. However, it requires the adjacency matrix as an input, and therefore, 
unscaleable for large graphs. 
In this paper, we propose a novel formulation for solving IM that addresses some of the challenges that we discussed so far. 


\section{Formulation of Activation informed IM (AIM)}
\label{sec:formulation}
The overall architecture of the proposed framework (\textsc{GraMeR}) is depicted in Fig. \ref{fig:1a}, and consists of three main modules. Each of these modules is explained in forthcoming sections. In this section, we describe the IM model considered in our work. It is reasonably different and generic relative to typical IM models explored in related recent work. 
In a social network, users propagate their views or opinions while simultaneously consuming and reacting to content created by friends, people, and organizations they follow. Thus, there are two ways of activation, namely {\em intrinsic} (content creation) and {\em spread of influence} (content spreading). 
Conventional IM problem solely considers influence activation and overlooks intrinsic activation. However, in practice, the role of content creators has gained significant importance due to massive digitization. Therefore, in this work, we are considering a generic formulation of IM that incorporates both types of activations and is referred to here as activation-informed influence maximization (AIM).

Enabled by the digital revolution, most users now are both content creators and content spreaders at the same time. This is probabilistically modeled through parameters $p_{s}$ and $p_{f}$ which represent the probability of intrinsic and influence activation, respectively. 
Intrinsic activation for a user $u$ is based on its own activities, and user $u$ is assumed to be directly influenced by its 1-hop neighbors. 
Therefore, the influence part of the probability for activation is comprised of the activation probabilities due to the 1-hop neighbors of user $u$. Thus, similar to the IC model, the probability of user $u$ being activated via influence of an user $v$ is written as:
\begin{equation}
    p_{uv} = w_{uv} p_{f}(u),
    \label{eq:0a}
\end{equation}
where the weights $w_{uv}$ $(0 \leq w_{uv} \leq 1)$ can be determined from the user interactions in a network. 
The described probabilistic formulation has similarities to the Friedkin-Johnsen social influence model for opinion change \cite{friendkin1999social}, where the authors recognize that the dynamics of opinion change are governed by two mechanisms: intrinsic opinion and influenced opinion. Furthermore, by assuming that the nodes are not lazy and are activated by either of the two mechanisms that we outline, we set $P_{f}(u) = (1 - P_{s}(u))$. This renders the overall IC probability between nodes $v$ and $u$ to be:
\begin{equation}
    p_{uv} = w_{uv}(1-p_{s}(u)).
    \label{eq:0b}
\end{equation}
Note that all the parameters discussed can be efficiently determined either by a maximum-likelihood-based approach or expectation-maximization (EM) approach as followed in \cite{saito2008prediction}. 
For example, in the Twitter network, the proportion of tweets by a user $i$ that are intrinsic in nature can quantify $P_{s}(u)$, while a particular weight $w_{uv}$ can be determined by the proportion of user $u’s$ retweets (or influenced activity) having their origin in the activity of user $v$ that user $u$ follows.


\section{Prediction of Candidate Nodes}
\label{sec:Prediction}
A fundamental aspect of our framework is the fact that only a certain fraction of nodes are likely to contribute to the AIM solution set. These nodes are referred to as \quotes{candidate nodes} in this work. Hence, it is desirable that instead of focusing on all nodes, one should attempt all computationally expensive predictions for the candidate nodes. This section describes a novel GNN based node classifier that leverages GNN based classifier to identify candidate nodes for AIM. This classifier precedes our primary DRL algorithm (GraMeR) as shown in Fig. \ref{fig:1a}, and thereby eases the computational burden. 

The task of identifying candidate nodes is framed as a binary node classification problem where the two classes denote \quotes{candidate} and \quotes{non-candidate} nodes, respectively. The ground truth labels can either be generated via standard greedy hill climbing algorithm or novel centrality metrics recently being proposed \cite{singh2019centrality, hussain2021influence}. This work uses Influence capacity metric as it is computationally easy to compute compared to other approaches. Influence capacity (IFC) is a novel centrality metric to identify influential nodes. The influence of any node depends on its neighbor connection (local influence) and its own location in the graph (global influence). Local influence of a node $u$ ($I_{L}(u)$) can be estimated as \cite{singh2019centrality}:
\begin{equation}
    I_{L}(u)= 1 + \sum_{v \in N(u)} P(u,v) + \sum_{v \in N(u)} \sum_{z \in N(v)} P(u,v)P(v,z),
    \label{eq:1}
\end{equation}
where, $P$ is the influence probability associated with links, and the operator $N(.)$ denotes the neighbors. Likewise, the global influence score ($I_{G}(u)$) can be expressed as:
\begin{equation}
    I_{G}(u)= k_{c}(u)\big(1+\frac{D(u)}{D_{N}}\big),
    \label{eq:2}
\end{equation}
where, $k_{c}(u)$ and $D(u)$ represent coreness score and degree of node $u$, respectively. Notation $D_{N}$ denotes maximum node degree in the graph. The overall influence capacity $I(u)$ of a node $u$ can then be written as:
\begin{equation}
    I(u)= \frac{I_{L}(u)}{\text{max}_{v \in V } I_{L}(v)}\times \frac{I_{G}(u)}{\text{max}_{v \in V } I_{G}(v)}.
    \label{eq:3}
\end{equation}

Nodes retaining extreme values of IFC are labeled as \quotes{candidate nodes} while the rest of the nodes are marked as \quotes{noncandidate}. The threshold of approx $20\%$ seems to work in our case and it is determined based on several experiments, where the candidacy of nodes is validated with the standard Greedy hill-climbing algorithm. The ground truths from IFC are then leveraged to train the node classifier without relying on a computationally exhaustive Greedy hill-climbing approach~\cite{Minutoli2019}. It is important to note that there are various other algorithms to generate groundtruth and one can take any such approach for training the node classifier.

The GNN based node classification model consists of two hidden layers with GraphSAGE as the message-passing algorithm. We selected GraphSAGE since the computation graph for any node $u$ only depends on the induced subgraph up to $k$-hop neighbors of $u$. This allows training/prediction across different graph sizes which is most desirable for GraMeR. The parameters of GNN are trained by minimizing the categorical cross-entropy loss function. The overall candidate node prediction task is summarized in Algorithm A.1 in the appendix. It is interesting to note that if this GNN classifier can provide a confidence interval around its class predictions as shown in \cite{munikoti2022general}, then the succeeding DRL engine can utilize that interval while searching for optimal seed nodes. This further strengthens the overall framework and is kept as future work.


\section{Meta Reinforcement Learner}
\begin{figure*}
\centering
	\includegraphics[width=0.9\textwidth]{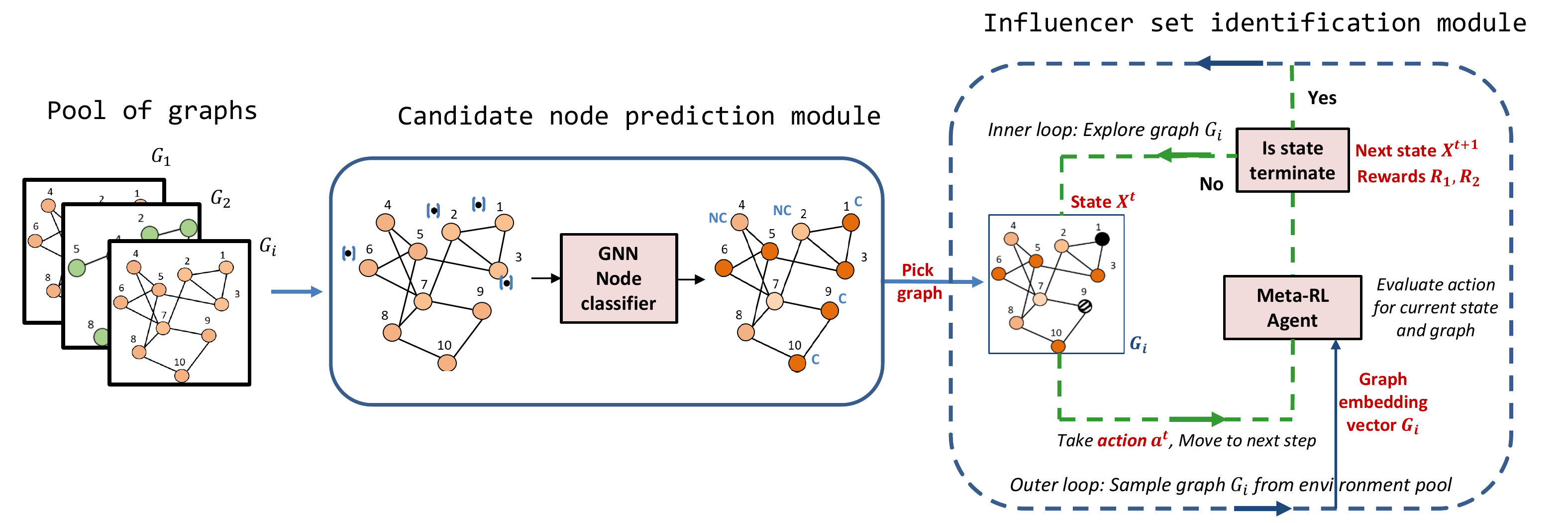}
	\caption{Basic architecture of GraMeR. Graphs of different types and sizes are provided to Candidate node prediction module for identifying candidate nodes (C: Candidate; NC: Non-candidate). The Influencer set identification module uses Meta DRL algorithm to train on the candidate nodes of a graph.}
	\label{fig:1a}
\end{figure*}

Graph Meta Reinforcement Learner (\textbf{GraMeR}) is a deep reinforcement learning (DRL) based framework that learns to identify optimal seed set for AIM. There are various novel aspects of our framework in terms of: (1) Meta learning to enable prediction across different graph types and sizes; (2) double Q learning to estimate sequence of seed nodes without solving a computationally intensive optimization problem at every test time and (3) single policy multi-objective reward formulation for systematic balancing of multiple AIM objectives. This section describes these novel aspects of GraMeR, and architecture is shown in Fig. \ref{fig:1a}. 

\subsection{Finite MDP}
The task of identifying an optimal set of AIM seed nodes is a sequential process where nodes are added one at a time. More importantly, the factors determining the selection of node at any step of the process solely depends on the last node added to the sequence (solution set). Thus, this process follows the Markov property and is therefore formulated as a Markov decision process (MDP). Further, at any step of the process, the action space is finite, i.e., a node has to be selected from a finite set of nodes. So, more precisely, the process can be termed as a finite MDP.
The key ingredients to define any finite MDP in the context of DRL are:\\
\textbf{State} represents the current solution set where nodes are appended in sequence to form the final AIM solution set. Thus, cardinality of the state keeps increasing with the process. Therefore, a state representing vector ($\mathbf{X^{t}}$) of fixed dimension is needed. $\mathbf{X^{t}}$ should characterize the state of the system at any time step $t$ in terms of nodes being selected. Therefore, $\mathbf{X^{t}}$ can be expressed as:
\begin{equation}
    \mathbf{X^{t}} = f(S^{t})
\end{equation}
where, $S^{t}$ is the partial AIM solution set at time $t$ and $f$ is the transformation operator. As nodes in the state are sampled from a graph, an appropriate choice for $f$ operator would be a Graph neural network based transformation which will be discussed in detail in the forthcoming sub-section. \newline
\textbf{Action} refers to the process of adding a new node $u$ to a partial solution set $S^{t}$.\newline
\textbf{Reward} quantifies the benefit of taking an action. There are two objectives in our AIM formulation which leads to two reward functions. The first reward ($R_{1}$) belongs to the marginal gain in influence spread when a particular node has been added to the solution set. The second reward ($R_{2}$) corresponds to the intrinsic probability of node being added to the solution set. The rewards can be written as:
\begin{equation}
\begin{split}
    R_{1}(X,a) & = \mathcal{I}(G,S \cup \{a\}) - \mathcal{I}(G,S) \\
    R_{2}(X,a) & = p_s(a)
    \end{split}
    \label{eq:7n}
\end{equation}
where, the operator $\mathcal{I}$ computes the influence spread of solution set $S$ in graph $G$ under the independent cascade model.
\newline
\textbf{Environment:} It's an agent world with which it interacts and comprises of everything outside the agent. Here, the environment is a graph. These interactions occur continually, i.e., agent selects node and the graph environment responds to those actions and present new situations to the agent.
\newline
\textbf{Policy:} The policy is a strategy or suggested actions that the DRL agent should take at every state of the environment so as to pursue the goal of the learning. It is a probability distribution over feasible nodes that could be added to partial solution set $S^{t}$ to move the state from $X^t$ to $X^{t+1}$. Hence, policy $\pi(a|X^{t})$ selects the node that yields highest cumulative reward at any arbitrary state $X^{t}$.
\newline
\textbf{Termination:} At every episode, the search starts with a random node from the candidate set, and the estimated nodes are appended to the partial solution set, one at a time ( one at each step of the episode). The episode is terminated when the cardinality of the solution set $S^{t}$ attains the search budget $b$.

\subsection{Algorithm}
GraMeR consists of three modules as illustrated in Fig. \ref{fig:1a}. The first module acts as an environment pool containing training graphs from different families and sizes. The second module provides a set of candidate nodes (described in Section 4) on which the AIM search algorithm will be implemented. Finally, the third module deals with the DRL agent. The process of training the agent starts by randomly selecting a graph $G_{i}$ from the pool and passing through second module to generate a candidate node set. Each training graph serves as an environment with which our agent interacts via MDP. An episode starts with a random node from the candidate set of the sampled graph and it continues until the budget is consumed. At each step of the episode, the agent will select the next node based on its current policy. Thereafter, the agent updates its current policy by training a Q network with a sampled batch of data from the buffer. Replay buffer stores the state, action and rewards from all the past steps of the process across episodes and environments (graphs), allowing Q network to exploit known information. Once the episode meets termination criterion, the agent samples a new graph and the training iterates until the policy converges. 

\textit{ Vanilla Q Learning:} The agent in GraMeR trains via double Q-learning as it is a discrete finite MDP. The vanilla Q-learning maximizes a cumulative reward of actions taken during the interactions of agent with environment \cite{watkins1992q}. 
Reward at future times depends on actions taken at current time.
The optimal value of an action (i.e., Q-value) corresponds to the optimal policy that maximizes the Q-value. Therefore, Q-value is iteratively updated according to Bellman equation as,
\begin{equation}
\begin{split}
    Q(X^{t},a^{t}) = Q(X^{t},a^{t}) + \\
    \theta * [r^{t} + \gamma\max_{a'}  Q(X^{t+1},a')-Q(X^{t},a')]
    \end{split}
    \label{eq:7}
\end{equation}
Q learning using Eq. (\ref{eq:7}) usually suffers from overestimation in practice due to the use of single estimator (Q-network) that determines the best action at next state with highest Q-value as well as the Q-value of that best action \cite{smith2006optimizer}. To avoid overestimation, double Q-leaning is proposed in \cite{hasselt2010double}, that uses two different estimators. One estimator (Local network $Q^{L}$) determines the best possible action for the next state and the other (target network $Q^{T}$) provides the Q-value of the selected action. The modified update equation of $Q^{L}(X^{t},a^{t})$ is,
\begin{equation}
\begin{split}
    Q^{L}(X^{t},a^{t}) + \theta * [r^{t} + 
    \gamma  Q^{T}(X^{t+1},a^{*})-Q^{L}(X^{t},a^{t})], \\
      a^{*}  = \argmax_{a'} Q^{L}(X^{t+1},a'),
    \end{split}
    \label{eq:8}
\end{equation}

and $\theta$ is the tuning parameter. Local network ($Q^{L}$) is trained at every step of the episode by sampling a batch of data from replay buffer. Mean squared error loss between the predicted Q value (i.e., $Q^{L}(X^{t},a^{t})$) and the desired Q value from the bellman equation ($r^{t} + \gamma  Q^{T}(X^{t+1},a^{*})$) is minimized to update the parameters of the local $Q^{L}$ network. While, the target network $Q^{T}$ is not explicitly trained at every step, it is continuously updated with the weights of entire $Q^{L}$ network after a certain number of episodes.

\textit{ Meta Q Learning:} A meta-learning attribute is introduced in the GraMeR to solve unseen tasks fast and efficiently. Here, the agent is expected to generalize to new graph types that have never been encountered during training. Typically, the Meta reinforcement learning approach contains two optimizer loops \cite{botvinick2019reinforcement}. The outer optimizer samples a new environment in every iteration and adjusts parameters that determine agent behavior. In the inner loop, the agent interacts with the environment and optimizes for maximum reward. As in most of the environments (such as mazes, self driving car, etc.), it is not feasible to obtain a representing vector for entire environment. Therefore, learning across environments is captured via an outer optimizer. However, in the case of AIM, the environment is a graph, and it can be represented very accurately by a single graph embedding vector \cite{zhang2018end}. Hence, we skip the outer optimizer and feed the entire environment (graph) information ( graph embedding vector) along with state and actions to the training algorithm. This enables Q learning to capture the variation of environment via a single optimizer.
Then the exact update equations of our agent turns out to be,
\begin{equation}
\begin{split}
    a^{*} = \argmax_{a^{'}} Q^{L}(X^{t+1},a',G_{i}) \\
    Q^{L}(X^{t},a^{t}, G_{i}) = Q^{L}(X^{t},a^{t}, G_{i}) + \\
    \theta * [r^{t} + \gamma  Q^{T}(X^{t+1},a^{*}, G_{i})-Q^{L}(X^{t},a^{t}, G_{i})]
    \end{split}
    \label{eq:9}
\end{equation}
where, $G_{i}$ corresponds to the sampled graph for the $i^{th}$ episode. It is worth noting that graphs changes with episodes but for a particular episode, a single graph is explored. 

\textit{GNN encoding:} Similar to the graph, we also need representing vectors for the state (partial solution set ) as well as action (node). In this regard, we leverage GraphSAGE \cite{hamilton2017inductive} to estimate node embeddings. Thereafter, the embedding vectors of nodes in the state $\mathbf{S^{t}}$ are aggregated via mean/max operation to obtain a single representing vector {$\mathbf{X^{t}}$ for the entire state. As action corresponds to a single node, it is represented by the corresponding node embedding vector. GraphSAGE is selected over conventional GCN \cite{kipf2016semi} due to the factor that the candidacy of a particular node to be a part of AIM solution set depends mostly on its sub-graph. Therefore, GraphSAGE being an indutive sub-graph based learning approach is an appropriate choice. Further, GraphSAGE does not demand entire graph information unlike GCN, which requires the entire adjacency matrix and thus does not scale well with the size of the graph. 

\textit{Multi-objective shaping:} Along with states and action, the reward function needs to be explicitly designed for AIM as it comprises of multiple objectives. The first objective is related to maximizing influence spread and the other goal corresponds to maximizing intrinsic probability of seed nodes. Therefore, GraMeR belongs to the category of multi-objective DRL \cite{nguyen2020multi}. We take a single policy approach to learn one optimal policy by combining the two objectives with a known preference weight. Precisely, at each step of the episode, rewards (Eq. (\ref{eq:7n})) are computed individually for each objective and accumulated in the buffer along with state and actions. Then, Q values of two objectives are combined using linear scalarization technique to generate a single Q value that is used to select an action. This is different than a typical approach of combining rewards into a single value and consequently learning single Q value and it is shown to be a stable and efficient \cite{van2013scalarized}. Algorithms 1 and 2 summarizes the entire GraMeR involving GNN based state/environment representation, meta learning across different graph types and multi-objective rewards shaping.

\begin{algorithm}
 \caption{ Pseudocode of GraMeR }
 \begin{algorithmic}[2]
 \renewcommand{\algorithmicrequire}{\textbf{Input:}}
 \renewcommand{\algorithmicensure}{\textbf{Output:}}
 \REQUIRE set of training graphs $G$, input node features $X$, set of candidate nodes $C$.
 \ENSURE  Trained GraMeR agent.
 \STATE Initialize: State $X \leftarrow \text{Max}$ $\{ h_{u} \forall u \in S \}$ 
 \STATE Initialize: Graph embedding to $G_{i}$ $\leftarrow$ $\text{Max} \{ h_{u} \forall u \in V \}$
 \STATE Initialize: Action $a$ corresponds to selection of node $u$ $\leftarrow$ $h_{u}$ where $h_{u}$ denotes node embedding from GraphSAGE.
 \STATE Initialize: Some arbitrary values to $Q_{1}(X,a)$, $Q_{2}(X,a)$  for all state-action pairs. 
  \FOR { Loop for each episode}
  \STATE Initialize: $X^{0}$ to initial state (random node from $C$)
    \FOR { Loop for each step of episode until state is terminal }
  \STATE  Select node $a^{*}$ from \textit{scalarized action selection strategy}
  \STATE  Append node $a$ to partial solution set $S$.
  \STATE  Obtain rewards $r1,r2$ from both objectives.
  \STATE  Move to next step $X^{'}$.
  \STATE  Update $Q^{L}$ of both objectives.
  \STATE  $Q^{L}_{1}(X,a, G_{i}) = Q^{L}_{1}(X,a, G_{i}) + \theta*[r + \gamma  Q^{T}_{1}(X,a^{*}, G_{i})-Q^{L}_{1}(X,a, G_{i})]$
  \STATE  $Q^{L}_{2}(X,a, G_{i}) = Q^{L}_{2}(X,a, G_{i}) + \theta*[r + \gamma  Q^{T}_{2}(X,a^{*}, G_{i})-Q^{L}_{2}(X,a, G_{i})]$
  \ENDFOR
  \ENDFOR
 \STATE update $Q^{L}_{1}$ and $Q^{L}_{2}$  parameters by minimizing cross entropy loss between target values from bellman equation and actual prediction from networks. 
  \STATE update $Q^{T}$ network by copying $Q^{L}$ parameters to $Q^{T}$ at every $U$ steps.
 \RETURN Trained models $Q^{L}_{1}$ and $Q^{L}_{2}$
 \end{algorithmic} 
 \end{algorithm}

\begin{algorithm}
 \caption{ Scalarized action selection}
 \begin{algorithmic}[3]
 \renewcommand{\algorithmicrequire}{\textbf{Input:}}
 \renewcommand{\algorithmicensure}{\textbf{Output:}}
 \REQUIRE Q values of both objectives for current state-action pair $(X,a)$. 
 \ENSURE action $a^{*}$. 
 \STATE Qnew = $W_{1}*Q_{1}(X,a) + W_{1}*Q_{2}(X,a)$
  \STATE Greedy action selection $$ a^{*}  = \left\{
	\begin{array}{ll}
		\argmax A_{a^{'}}  & \text{probability} \hspace{0.1cm} 1-\epsilon \\
	    \text{random selection from set C}    & \text{probability} \hspace{0.1cm} \epsilon 
	\end{array}
\right.$$
\RETURN action $a^{*}$
\end{algorithmic} 
\end{algorithm}
 
\section{Experimental Results}
\label{sec:Experiments}
This section validates the proposed framework against modified greedy hill-climbing algorithm (MGHC) and S2VDQN. GraMeR offers improved performance with more flexibility while being orders of magnitude faster. The performance is examined on standard networks, namely Barabasi Albert, Power law cluster, stochastic block models. As per the architecture, $1^{st}$ phase of the local Q-network ($Q^{L}$) comprises of $3$ GNN layers for generating node embeddings. The number of neurons in these layers are $64$, $32$ and $16$, respectively. This phase generates state, action and graph embedding vectors which are  concatenated into a vector and passed through a regression phase of the $Q^{L}$ network. Regression phase consists of $2$ feedforward layers with $16$ and $1$ neurons respectively. All other settings have been discussed in the appendix.

\subsection{Baselines}
We have first validated the performance of the proposed candidate node prediction model with Greedy Hill-climbing (GHC) algorithm that selects seed nodes based on the marginal gain in influence spread. Then, GraMeR (core module) is compared with modified greedy Hill-climbing (MGHC) algorithm \cite{sathanur2018exploring} that sample nodes based on intrinsic probability and select seed nodes which provide high gain in spread. Although this algorithm strives for two objectives (i.e, high intrinsic probability and high marginal gain in influence spread) of AIM, there is no inbuilt mechanism to control their priorities. In fact, none of the recent data-driven work addresses this type of multi-objective formulation of IM \cite{dai2017learning, manchanda2020gcomb}. In contrast, we have systematically incorporated the multi-objective formulation with a controlled weighing factor $\alpha$. Furthermore, the baseline also includes modified S2V-DQN (MS2V-DQN) \cite{dai2017learning} that combines RL with graph representation. The baselines implementation is further discussed in the Appendix.

\begin{figure}
\centering
	\includegraphics[width=0.92\linewidth]{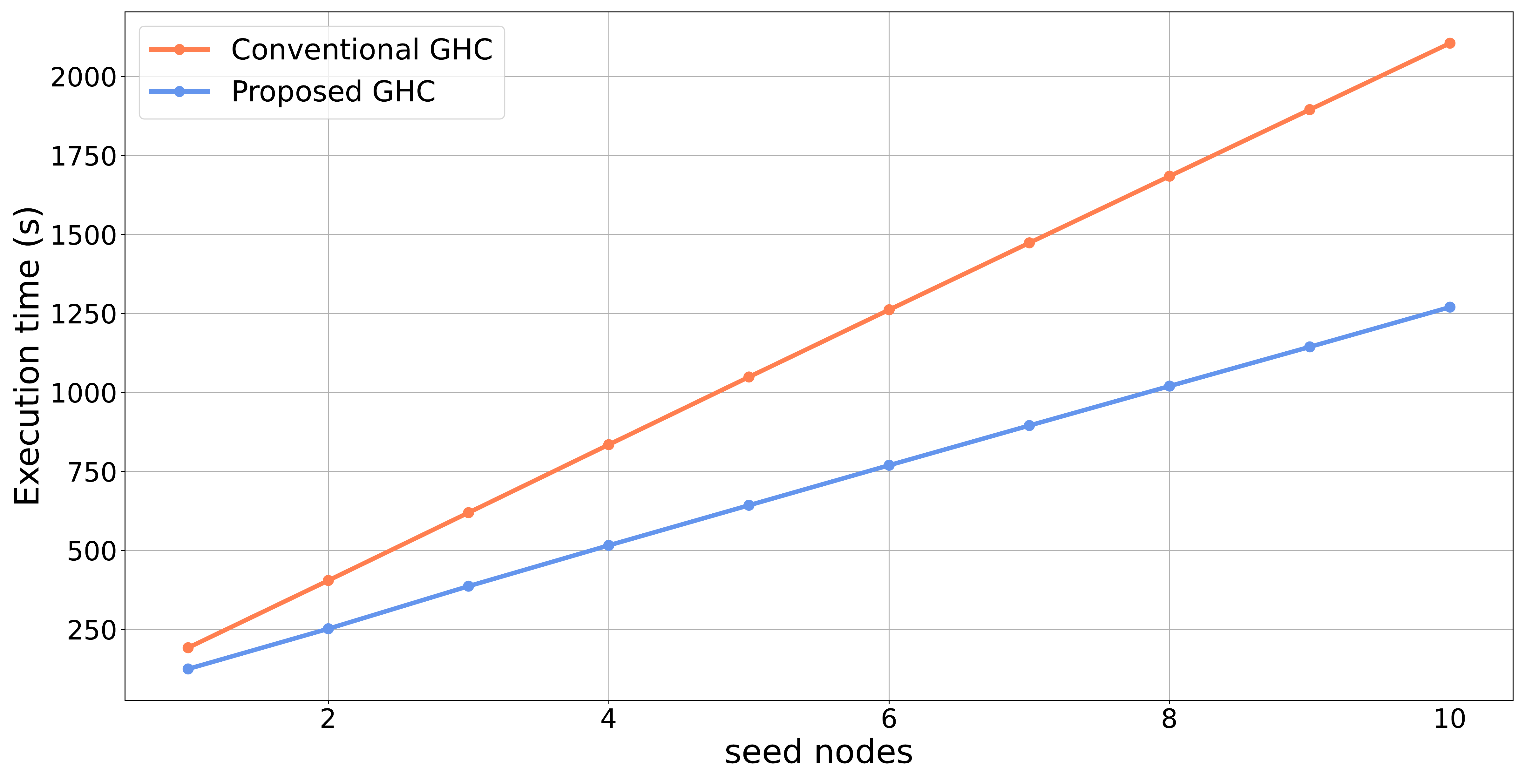}
	\caption{Time in identifying seed nodes with and without candidate nodes}
	\label{fig:3b}
\end{figure}

\begin{figure*}
\centering
	\includegraphics[width=0.8\textwidth]{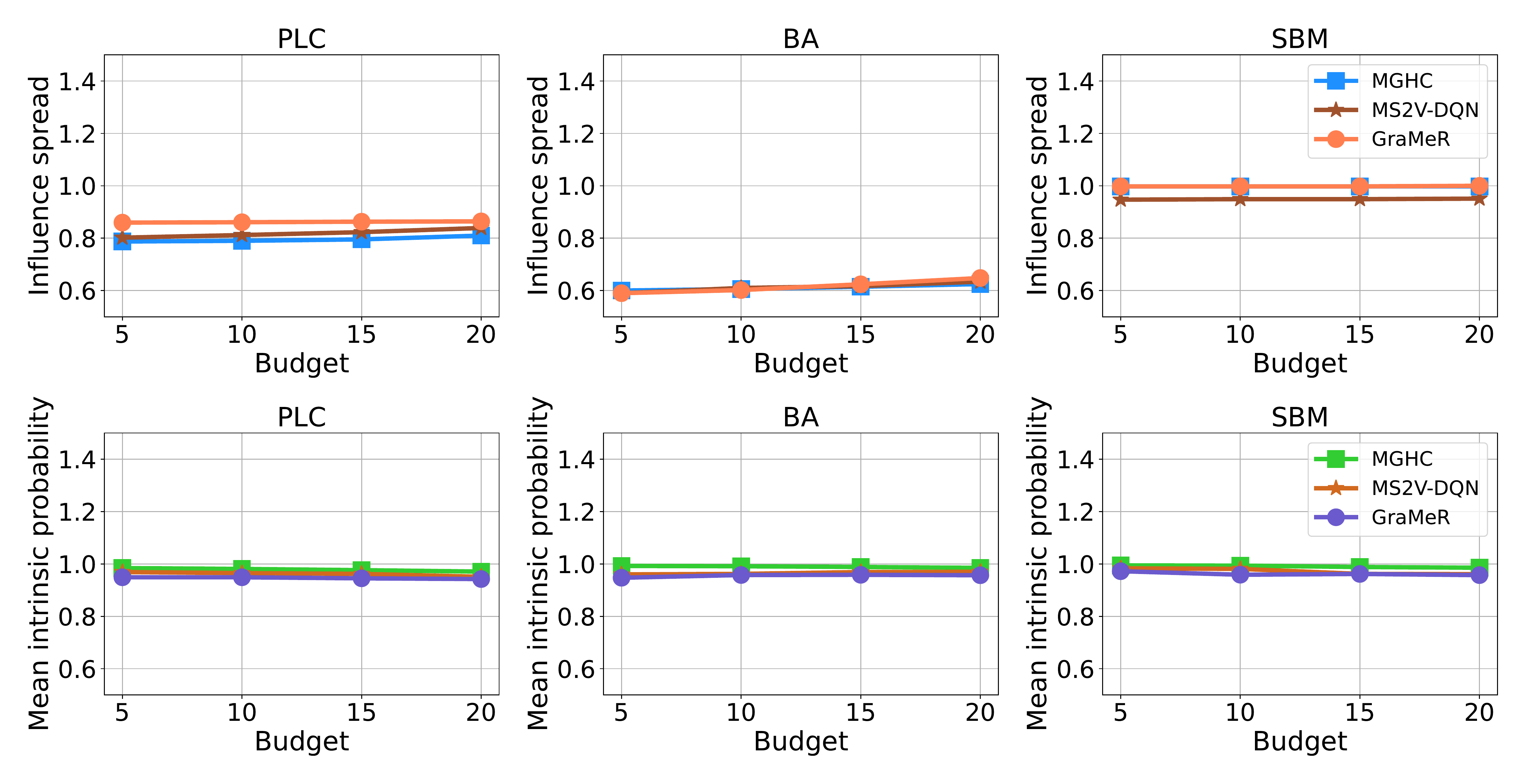}
	\caption{Spread/Intrinsic probability vs Budget. PLC: Power law cluster; BA- Barabasi-Albert; SBM-Stochastic block model}
	\label{fig:2}
\end{figure*}

\begin{figure*}
\centering
	\includegraphics[width=0.7\textwidth]{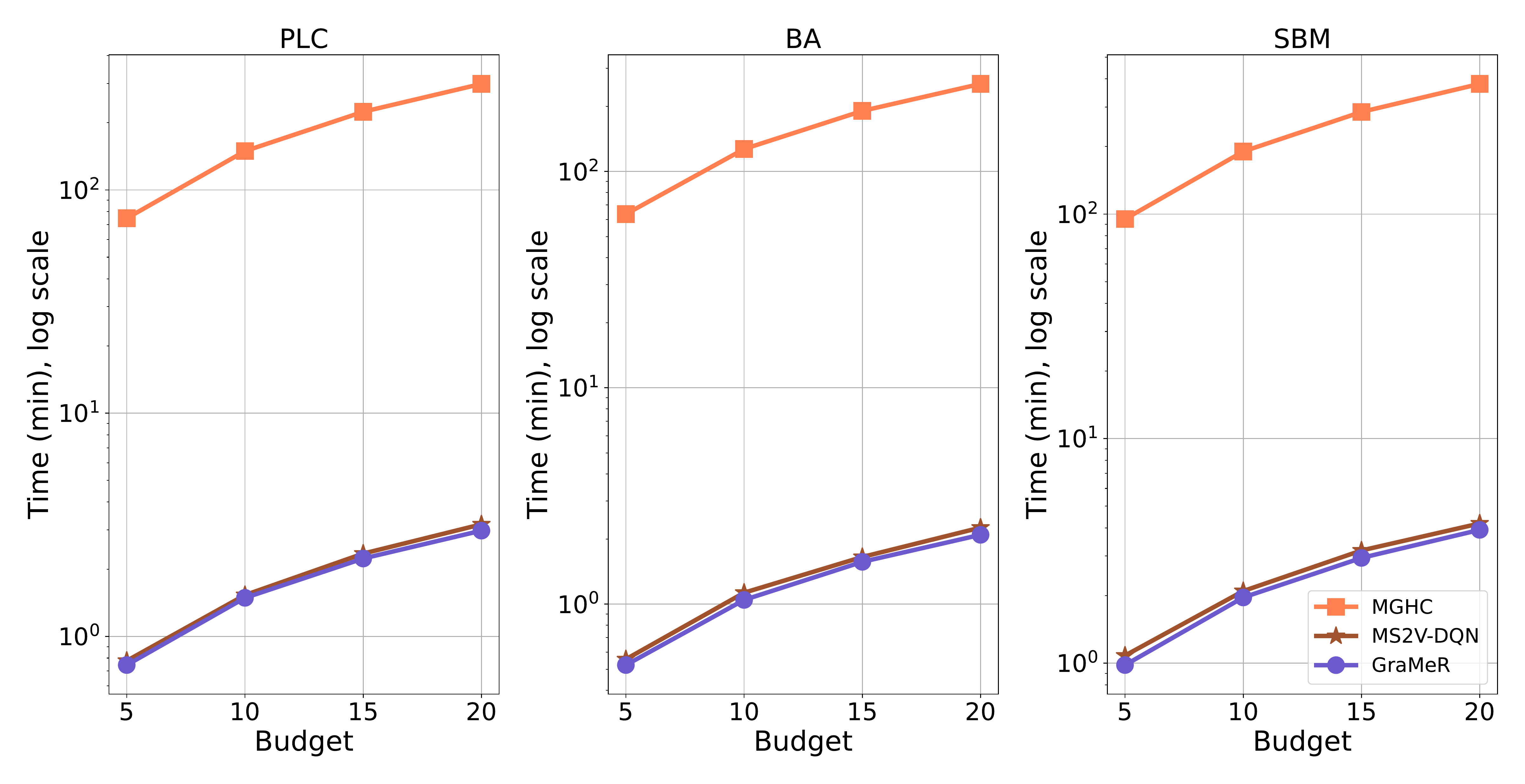}
	\caption{Running time vs Budget. PLC: Power law cluster; BA- Barabasi-Albert; SBM-Stochastic block model}
	\label{fig:3}
\end{figure*}

\begin{figure*}
\centering
	\includegraphics[width=0.85\textwidth]{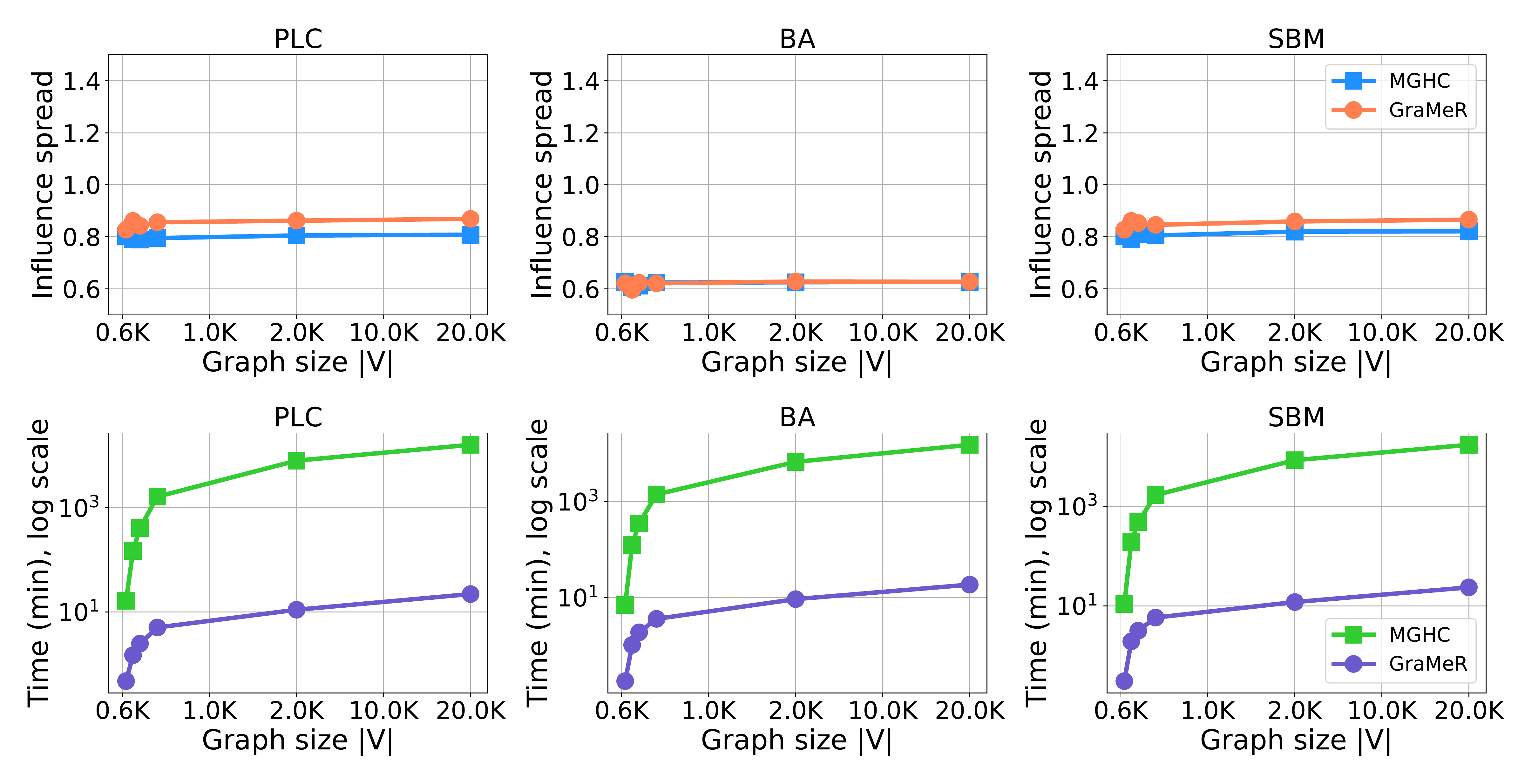}
	\caption{Mean Spread and Running time vs graph size for three graph types}
	\label{fig:4}
\end{figure*}

\begin{figure*}
\centering
	\includegraphics[width=0.7\textwidth]{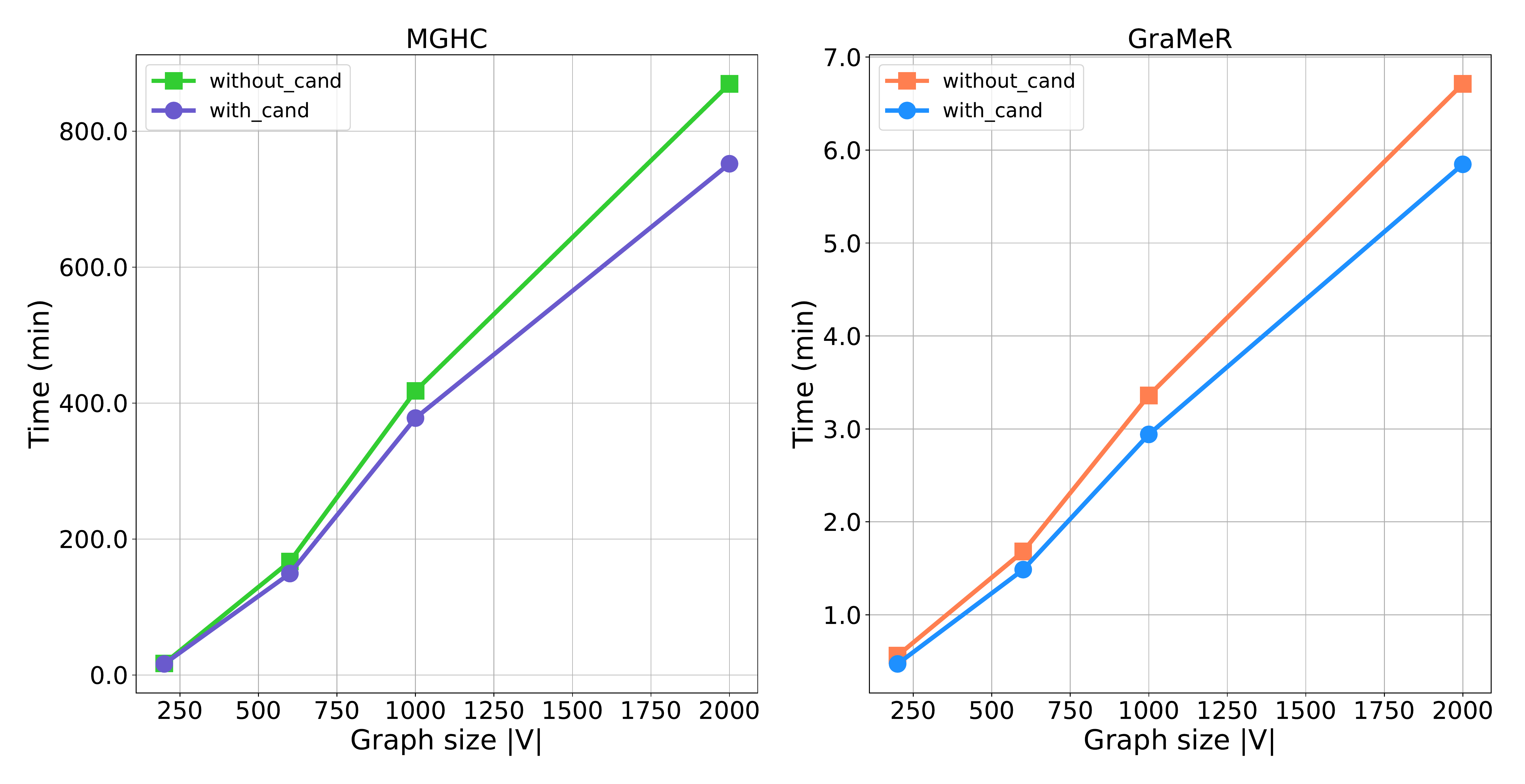}
	\caption{Running time vs graph size of GraMeR and MGHC for PLC graph}
	\label{fig:5}
\end{figure*}

\subsection{Results}
This section demonstrates the performance of proposed framework against baselines via performance metrics and execution times. The evaluation is repeated $100$ times and average scores are reported for each test scenario. The source code can be found in the following anonymized github link https://anonymous.4open.science/r/InfluenceMaximization-Deep-QLearning-B20B.

\subsubsection{Performance of Candidate node predictor}
The candidate node prediction module is trained on $6$ graphs of varying dimension ($200$ to $400$ nodes) and type (BA, PLC, BP). The ground truth value of node classes is obtained using influence capacity metric as discussed in Eq.(\ref{eq:3}). The model is evaluated on $3$ test graphs of $1000$ nodes, one from each graph family. The mean classification accuracy in detecting candidate and non-candidate nodes is around $96.24$ \% with a recall score (class-1 accuracy) of $97.95$ \%. Further, it has been shown via Fig. \ref{fig:3b} that there is a noticeable reduction of $40$\% in algorithm running time compared to the greedy hill-climbing approach. This is due to a significant reduction in search space which can immensely speedup the training of GraMeR agent in later part of the learning pipeline.

\subsubsection{Accuracy of GraMeR}
The performance of GraMeR for activation aware influence maximization can be evaluated via two metrics namely influence spread and node intrinsic probability. Expected influence spread provides the mean number of nodes getting influenced in the network if seed nodes in the solution set are activated with corresponding intrinsic probabilities and information is spread via IC diffusion model. Influence spread is normalized between $0$ to $1$ for comparison across different graph sizes. The second metric (i.e., intrinsic probability) represents the probability of seeds nodes being activated on their own. The mean probabilities for all the seed nodes in the estimated solution set are reported.

Fig. \ref{fig:2} compares expected influence spread and mean intrinsic probability across different graph types and budget. It can be observed that the proposed GraMeR is at par with the baseline MGHC with much lower computational effort as demonstrated in the next subsection. Specifically, GraMeR consistently outperforms MGHC and MS2V-DQN in terms of spread whereas the reverse phenomenon can be seen in case of intrinsic probability. This is because the MGHC and MS2V-DQN does not have a mechanism to balance between influence spread and intrinsic probability. Therefore, they always provides the seed nodes having high intrinsic probability but sub-optimal in terms of influence spread. The scale of the plots is kept same for a fair comparison across different graph types. Further, it can be seen that for a similar graph size ($600$ nodes) and diffusion model, the influence spread is maximum in case of SBM graph and minimum in BA. This is due to the fact that SBM have community structures where subsets of nodes are connected with each other through a large link densities. AIM led to the selection of seed nodes from different clusters which results in a high influence spread compared to BA graph that misses such clusters.

\subsubsection{Computational gain of GraMeR}
One of the key advantages of the proposed approach is the computational efficiency which is demonstrated via algorithm running times i.e. wall clock time. Fig. \ref{fig:3} depicts the running times across different budgets and graphs. It can be inferred that as the budget increases, the search time increases which is very intuitive. However, the increase in time is very sharp (linear for the experimented budgets) in the case of baseline MGHC, while it is almost constant for the proposed GraMeR. This is because the trained deep Q networks in GraMeR computes the solution set via forward propogation which mainly involves matrix operations. On the other hand, the MGHC searches for increasing number of nodes as the budget increases which eventually demands the computation of influence spread for large number of sets.
Fig. \ref{fig:3} also illustrates that search algorithms for AIM are fastest in BA and slowest in SBM. This observation could be attributed to the high clustering coefficient in SBM leading to a large time cost associated with shifting from one cluster to an other while searching for an optimum seed set.

\subsubsection{Generalizability of GraMeR}
The core theme of our framework reinforces the property of generalizability as there are various types of real-world and synthetic networks that exist in the literature. Many of these graph types have only slight variation in their topological property. Thus, a separate DRL model to identify AIM seed nodes for each of these graph types is not needed rather, a meta learning can serve the purpose by learning across different environments (graph types). This fact is demonstrated by training on two graph types (i.e., PLC and BA) while validating on all the three graph families. The performance in terms of influence spread, intrinsic probability and running time is consistent across all three graphs as shown in Figs. \ref{fig:2} to \ref{fig:5}.   

\subsubsection{Scalability of GraMeR}
Scalability of GraMeR is summarized in Fig. \ref{fig:4}, which presents the compute time for different graph sizes. Here, the model is trained on $8$ graphs ($4$ from PLC and BA) each having $400$ nodes. Then, it is tested on graphs of sizes varying from $200$ to $20000$. It can be inferred that GraMeR outperforms the baseline without any significant impact on computational effort. Specifically, running time for MGHC scales by $1500$ for $10x$ increase in graph size whereas, it remains almost constant for GraMeR. Further, to reinforce the scalability benefit, the training of GraMeR is carried for a fixed budget of $10$ but prediction is carried out for multiple budgets from $5$ to $20$.
The scalability can further be enhanced through distributed computing as shown by \cite{minutoli2019fast}. This work will address scaling to larger graphs on a wide variety of platforms (GPUs), which we plan to pursue in our future work.

\subsubsection{Ablation study}
The proposed GraMeR has computational supremacy over conventional methods because of two factors: (1) Deep Q networks based meta reinforcement learning approach to identify AIM solution set; and (2) candidate node predictor that reduces the search space. 
Fig. \ref{fig:5} depicts the ablation study results where running time is monitored for GraMeR and MGHC with and without the candidate node prediction module. The budget is fixed as $10$ and prediction is done for the PLC graph across different sizes. It can be seen that the time gap between the two cases increases with the graph size with nearly $1$ minute for GraMeR and $110$ minute for MGHC. This gap will further increase as the network size grows. Further, apart from prediction, noticeable gap is also seen in training time. This demonstrates the importance of the node prediction module in GraMeR.

\section{Conclusions}
We presented a GNN fused meta reinforcement learning framework (GraMeR) for identifying influential nodes in a network. Firstly, the search space of IM is reduced via GNN based candidate node predictor. Then deep Q learning is employed to learn to identify IM seed nodes with GNN as environment encoders. The unique aspects of GraMer lies in its computational efficiency and generalizability. Future directions of research include extension of GraMeR for uncertain networks suited for financial/manpower constraint applications.   

\bibliographystyle{IEEEtran}
\bibliography{reference}




\appendix
\section{Appendix}
\subsection{Datasets}
The performance of GraMeR is examined on standard networks. The description of datasets are as follows:

\begin{enumerate}
    \item \textbf{Barabasi Albert (BA):} Graphs which are grown by attaching new nodes each with fixed number of edges that are preferentially attached to existing nodes with high degree. Several natural and real-world networks, including the Internet, citation networks, and some social networks are learned to follow this model. 
    \item \textbf{Power law cluster (PLC):} Graphs which exhibit both power law degree distribution and clusters, and many real-world networks manifest these properties. As shown in \cite{holme2002growing}, one can construct a PLC graph by following a process of preferential attachment but in some fraction of cases (p), a new node $N_{y}$ connects to a random selection of the neighbors of the node to which $N_{y}$ last connected.
  \item \textbf{Stochastic-Block model (SBM):} A Graph generated with k clusters, and a $k \times k$ (symmetric) matrix $P$ of probabilities, where $P_{i,j}$ is the probability that a pair of nodes $(a,b)$ will be joined by a link if $a$ is in cluster $i$ and $b$ is in cluster $j$.
    \end{enumerate} 
    
\subsection{Algorithm of candidate node prediction}
\begin{algorithm}[h!]
 \caption{ Training of Candidate node prediction module}
 \begin{algorithmic}[1]
 \renewcommand{\algorithmicrequire}{\textbf{Input:}}
 \renewcommand{\algorithmicensure}{\textbf{Output:}}
 \REQUIRE set of training graphs, input node features $X_{u}$, node labels $Y_{u} (\text{candidate or non candidate}) \hspace{0.1cm}  \forall \hspace{0.1cm}  u  \epsilon V$
 \ENSURE  Trained GNN
 \STATE Initialize: $h_{v}^{0}=X_{v}$  $\forall \hspace{0.1cm} v  \epsilon V$ 
  \FOR { layer $l=1$ to $3$ }
    \FOR { node $u=1$ to $u=V$ }
  \STATE  $h_{N(u)}^{l}$ = Attention($ h_{k}^{l-1} $) 
  $\forall \hspace{0.1cm} k \epsilon N(u) $ ;
  \STATE  $h_{u}^{l}=$ Relu($[h_{u}^{l-1}||h_{N(u)}^{l}]$)   $\forall \hspace{0.1cm} u  \epsilon V$ ;
  \ENDFOR
  \ENDFOR
 \STATE z = mean$(h_{u}^{l})$ $\forall \hspace{0.1cm} u  \epsilon V$ 
 \STATE $\hat{Y}_{u}$ = Linear($h_{u}^{l}|| z $)
 \STATE update GNN model by minimizing cross entropy loss between $\hat{Y}_{u}$ and $Y_{u}$
 \RETURN Trained GNN model
 \end{algorithmic} 
 \end{algorithm}

\subsection{Experimental setup}
All training and evaluation experiments are performed on a system with Intel i9-4820K processor running at 3.70GHz with 8 cores, having 1 Nvidia RTX 2080 Ti GPU with 12 GB memory, and 64 GB RAM. Evaluation is repeated $10$ times and average of metrics are reported.

\subsubsection{Evaluation settings}
We first created a training pool by randomly generating $15$ graphs from each graph family (Barabasi-Albert, Power-law cluster, Stochastic block model) with the NetworkX library. Ten graphs of each category are kept for training, and five for testing. The parameters of GNN based candidate node predictor classifier are:
Depth (\# of node embedding module): 2; \# of neurons in $2$ layers: 64,32; \# of MLP layers: 3; \# neurons in MLP layer: 12,8,1; Aggregator function: Mean; Activation function: ReLU (except last layer with sigmoid). The training is carried out via Adam optimizer with a learning rate of $0.0001$. The output of this module is a masked graph with candidate and non-candidate nodes.

The training of GraMeR is carried out on the masked graphs via double Q learning. Since states and action are represented via node embeddings, parameters of the $1^{st}$ phase of the GNN model are: Depth (\# of node embedding module): 3; \# of neurons in $3$ layers: 64, 32, 16. Aggregator function: Max. The parameters of $2^{nd}$ phase (i.e., regression) are: \# of MLP layers: 3; \# neurons in MLP layer: 12,8,1; Activation function: ReLU (except last layer with linear). Moreover, the settings specific to DRL are: Replay buffer size: 10000; batch size: 64; gamma: $0.99$; Learning rate: $0.0008$; target network ($Q^{T}$) update frequency: $10$; exploration decay rate: $0.996$. Mean squared error is used as a loss function to update $Q$ network weights with Adam optimizer.
Further, validation is conducted after each step of training, where the mean of reward functions on validation graphs is monitored.
The model is trained until rewards are saturated, which takes around $1000$ episodes in our case. Thereafter, the best model is evaluated on $15$ test graphs, $5$ for each graph family. All training and validation is carried out in PyTorch with the support of Deep Graph Library.

\subsubsection{Baseline details}
As a first baseline for AIM seed nodes, we have implemented a modified greed hill-climbing algorithm (MGHC)  proposed in \cite{sathanur2018exploring}. Let $S^{t}$ be the partial solution set of influential nodes at step $t$. The classical greedy hill-climbing optimizer expands the
set to size ($t + 1$) by polling each of the nodes not in $S^{t}$ and augmenting those
nodes, one at a time to form the set $S^{t} \cup \{u\}$ and looking for the best marginal
gain in terms of activations (no. of nodes influenced in the entire graph). At each such step $t$, instead of activating every single node in $S^{t} \cup \{u\}$  and then computing activations according to
the desired diffusion model, each node in the set $S^{t} \cup \{u\}$ is activated probabilistically with the corresponding node intrinsic probability thereby simulate the intrinsic activation process.
The modified part is depicted in line $9$ of Algorithm 1 in App A. Given the probabilistic
nature of the algorithm, the overall activation numbers are obtained by running the
diffusion model in a Monte Carlo fashion that invokes $M$ independent trials of
randomized graphs.







\end{document}